\documentclass[preprint]{article}

\usepackage{graphicx}
\usepackage{amsmath}
\usepackage{amssymb}
\usepackage{tikz}
\usepackage{natbib}

\usepackage[preprint]{neurips_2024}

\usepackage[utf8]{inputenc} 
\usepackage[T1]{fontenc}    
\usepackage{hyperref}       
\usepackage{url}            
\usepackage{booktabs}       
\usepackage{amsfonts}       
\usepackage{nicefrac}       
\usepackage{microtype}      
\usepackage{xcolor}         

\title{In-Context Learning of Physical Properties:
Few-Shot Adaptation to Out-of-Distribution
Molecular Graphs}

\author{%
  Grzegorz Kaszuba \\
  IDEAS NCBR, Poland\\
  \texttt{grzegorz.kaszuba@ideas-ncbr.pl} \\
  \And
  Amirhossein D. Naghdi \\
  IDEAS NCBR, Poland\\
  \texttt{amirhossein.naghdidorabati@ideas-ncbr.pl} \\
  \AND
  Dario Massa \\
  IDEAS NCBR, Poland\\
  \texttt{dario.massa@ideas-ncbr.pl} \\
  \And
  Stefanos Papanikolaou \\
  NOMATEN CoE, Poland\\
  \texttt{stefanos.papanikolaou@ncbj.gov.pl} \\
  \And
  Andrzej Jaszkiewicz\\
  Technical University of Poznan, Poland \\
  \texttt{andrzej.jaszkiewicz@cs.put.poznan.pl} \\
  \And
  Piotr Sankowski\\
  IDEAS NCBR, Poland \\
  \texttt{piotr.sankowski@ideas-ncbr.pl} \\
}

\begin{document}

\maketitle

\begin{abstract}

Large language models manifest the ability of few-shot adaptation to a sequence
of provided examples. This behavior, known as in-context learning, allows for performing nontrivial machine learning tasks during 
inference only. In this work, we 
address the question: can we leverage in-context learning to predict 
out-of-distribution materials properties? 
However, this would not be possible for structure property prediction tasks unless an effective method is found to pass atomic-level geometric features to the transformer model. To address this problem, we employ a compound model in which GPT-2 acts on the output of geometry-aware graph neural networks to adapt in-context information. 
To demonstrate our model's capabilities, we partition the QM9 dataset into sequences of molecules that share a common substructure and use them for in-context learning. This approach significantly improves the 
performance of the model on out-of-distribution examples, surpassing the one of general graph neural network models. 

\end{abstract}

\section{Introduction}

In-context learning refers to few-shot adaptation to a sequence
of provided examples. The ability of large language models, such as GPT-3 (\citep{brown2020language}), to perform in-context learning presents 
significant potential for application in scientific fields, particularly in material science and chemistry. In-context learning examples in text data such as: \vspace{0.3cm}

\centerline{$\underbrace{ \textrm{Mexico} \rightarrow \textrm{America}, \textrm{Italy} \rightarrow \textrm{Europe}, \textrm{Nigeria} \rightarrow \textrm{Africa}, \textrm{China}}_\textrm{Text Prompt}$  $\rightarrow$ $\underbrace{\textrm{Asia}}_\textrm{Prediction}$}

were also studied for chemistry-related prompts.
This includes evaluating a language model to predict molecule name, property, reaction 
prediction, etc., based on simplified molecular-input line-entry system (SMILES) in-context prompt inputs 
(\cite{guo2023large}) or using in-context learning for Bayesian optimization of catalysts (\cite{ramos2023bayesian}).

Models capable of generalization to novel compounds are highly valuable as conventional deep learning methods often 
struggle with out-of-distribution (OOD) 
predictions, creating a bottleneck in the discovery of functional drugs and materials. On the other hand, models capable of handling 
OOD data can significantly improve the screening of novel candidate materials. For instance, the GNoME model (\cite{Merchant2023}) led to the 
discovery of 2.2 million stable materials, 736 of which have already been experimentally validated. Another state-of-the-art deep 
learning model, MatterSim (\cite{yang2024mattersim}), demonstrated strong capabilities in making accurate predictions 
on OOD data and performed well in material discovery tasks, identifying 16,399 stable materials only within binary chemical systems. 
Finally, equivariant message passing neural networks, such as MACE (\cite{batatia2023mace}), have shown strong 
performance on OOD data (\cite{batatia2024foundation}). This capability is particularly beneficial for applications like materials screening and discovery. 

In our study, we explore the potential of in-context learning methods for predicting OOD physical properties based on atomic-level geometric features.
\begin{figure}[b]
    \centering
  \centering
  \includegraphics[width=1\textwidth]{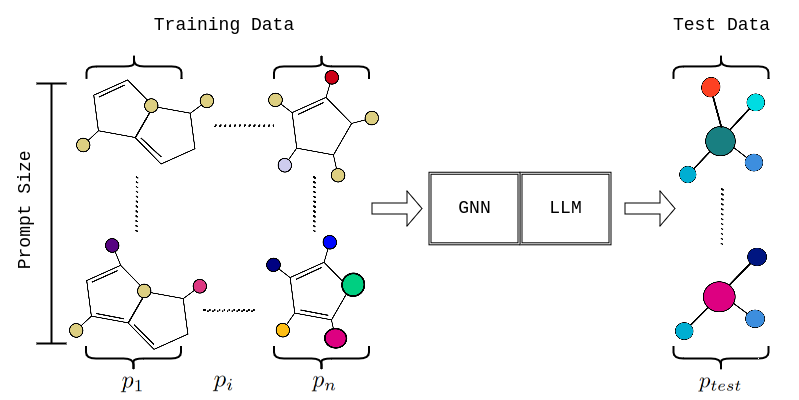}
  \caption{Schematic of partitioned dataset. Each prompt \( p_{i} \in \mathcal{P} \)  is created by dividing the dataset into molecule sequences that share a common substructure. The test data is chosen to ensure it has no overlap with the training set distribution.}
  \label{toc}
\end{figure}
While advanced deep neural network models for 
material discovery, which can handle OOD 
compounds to a certain degree, exist in the 
literature, the in-context learning method 
for this task is not well studied. (\cite{garg2023transformers}) explored 
well-defined problems to determine if 
transformers can learn a class of functions $F$ in-context and clarify the extent of this learning. They found that transformers are capable of 
in-context learning for simple function classes such as linear functions, as well as 
more 
complex ones like 3-sparse linear functions, two-layer neural networks, and decision trees. During each inference, the model is supposed to fit a function \( f \) from class \( \mathcal{F} \), when given a series of input-output pairs. Therefore, the variance of outputs has two sources: it depends of individual input for each pair, as well as the parameters of \( f \), which the model implicitly infers from the entire context.
This advanced capability of transformers opens up the possibility to equip them with 
real-world scientific data, such as 
incorporating geometric features of molecules 
and crystals to predict their structural properties.  


To advance our understanding of in-context learning for material discovery, we used this capability of transformers on QM9, a molecular property prediction benchmark \cite{wu2018moleculenet}. The function we predict is absolute-zero atomization energy \( \mathcal{U}_{0} \). As interactions that give rise to physical properties are immensely composite, we conceptualize that the estimator of \( \mathcal{U}_{0} \) shall be a function class \( \mathcal{F} \), demanding slightly different treatment \( f \) in different cases - namely, it will be predicted based on in-context examples of structures that share similar structural motifs. We therefore create prompts \( p \) of molecular graphs (and their respective labels), such that graphs have a common subgraph. Our main contributions are as follows: 

\begin{itemize}
\item Transformers can learn in-context from graph representations -- with the right curriculum transformers can be trained to perform in-context generalization on appropriate graph encodings.
\item The final model can generalize to OOD graph features -- the OOD test is performed on structures that have not been seen previously by the model. 
\item Formulating a strategy to utilize QM9 data for in-context learning -- we created a methodology aimed for benchmarking in-context learning models on molecular modeling tasks. 
\end{itemize}

The prediction of molecular properties, either quantitative or categorical, opens new pathways in numerous domains, e.g. in material, drug or protein discovery. The data-efficient adaptability of in-context learning models may enhance the exploration of novel material properties by effectively utilizing newly obtained samples to expand the model of knowledge. 

\section{Related work}
\label{related_work}


\paragraph{Learning representations of physical structures with Neural Networks.} 
Within the framework of Materials Informatics (\cite{ramakrishna2019materials,ward2017atomistic}), the search for an efficient definition of descriptors, capable of grasping molecules and crystal properties, stands as a key challenge which has been addressed in many different ways. Representations involving fixed length feature vectors encoding the physico-chemical properties of the systems \cite{isayev2017universal,xue2016accelerated}, as well as rotationally and translationally invariant functions of the atomic coordinates \cite{de2016comparing,behler2011atom}, have been widely used. However, the implementation of Neural Networks in the field of atomic physics and materials science has allowed for flexible and direct representation learning from the data, beyond the narrow scope of system-specific solutions offered by ad-hoc modelling of feature vectors. 
Differently from Convolutional Neural Networks based approaches (\cite{zheng2018machine,ryczko2018convolutional}), Graph Neural Networks (GNNs)  (\cite{scarselli2008graph,kipf2016semi}) are characterized by the possibility of directly relating to the structure and geometry of the atomistic data they are applied to, as the latter can naturally be encoded in the form of graph-structured data (\cite{duvenaud2015convolutional,back2019convolutional,xie2018crystal,chen2019graph}), mapping atoms to nodes and chemical bonds to edges. Early approaches to the problem of graph-networks for molecular data, like SchNet (\cite{schutt2017schnet}), consider edge lengths from interatomic distances, while further including edge features. An important aspect to consider during inference of molecular graphs is the geometry: in DimeNet (\cite{gasteiger2020directional}), lenghts and angles are encoded with radial and Bessel basis functions to then utilize in an edge-gated message passing operation. 
Directionality is also included in other approaches, including GemNet (\cite{gasteiger2021gemnet}), ALIGNN (\cite{choudhary2021atomistic}) and MXMNet (\cite{zhang2020molecular}). The latter, Multiplex Molecular Graph Neural Network, further considers an extra interaction scheme through global message passing to capture the different nature of covalent and non-covalent interactions, and has been implemented in the present work. 
\paragraph{Transformers and In-context learning for property prediction}
Concerning transformers (\cite{vaswani2017attention}), one of the examples for regression of molecular properties is represented by Mol-BERT: it involves pretrained BERT models to learn both structural and contextual molecular information (\cite{li2021mol}), and demonstrated superior classification performance compared to graph-based models. Other interesting methods in this context include ChemBERTa (\cite{chithrananda2020chemberta}), ChemBERTa-2 (\cite{ahmad2022chemberta}) and SMILES-BERT (\cite{wang2019smiles}). 

The emerging in-context learning capabilities of transformer language models enables pretrained models to swiftly adapt to novel tasks by leveraging a small set of demonstration examples during inference, avoiding the need for parameter updates (\cite{brown2020language}). 
\cite{chan2022data} underlined how both the transformer architecture and a \textit{naturalistic} training data distribution, characterized by burstiness and rarely occurring classes, play a critical synergistic role in the emergence of in-context-learning.\cite{huang2024prodigy} introduce Pretraining Over Diverse In-Context Graph Systems (PRODIGY), a novel framework allowing pretrained models to directly perform downstream tasks through in-context learning, without the need of adaptation to different tasks via fine-tuning, and therefore generalizing across different tasks within the same graphs as well as across different graphs.

Alongside the many contributions in recent years regarding in-context-learning (\cite{xie2021explanation,rong2021extrapolating,olsson2022context,min2021metaicl,min2021noisy,liu2021makes,lampinen2022can}), only few have, to our knowledge, focused on tasks related to materials science or chemistry. In particular, (\cite{ramos2023bayesian}) worked on an uncertainty equipped prompting system for in-context-learning with frozen LLMs, enabling Bayesian optimization for catalyst or molecule optimization using natural language prompts, eliminating the need for training or simulation. In another work, \cite{guo2023can} establish a comprehensive benchmark consisting of eight molecular tasks, involving property prediction, text-based molecular design, molecule captioning and selection tasks, and evaluate five well known LLMs, also considering different prompting and retrieval methods, varying the number of contextual examples in each task; their findings stress the importance of the quality and quantity of in-context examples in prompting performance, which is better than zero-shot prompting in all tasks. In the research field for drug design, \cite{edwards2023synergpt} deploy in-context learning techniques to uncover new drug synergy relationships for specific cancer cells targets, equipping them with a genetic algorithm to optimize model prompts and synergy candidates selection.

\section{Methods}
\label{Method}

\begin{figure}[h] 
\centering
\includegraphics[width=0.35\textwidth]{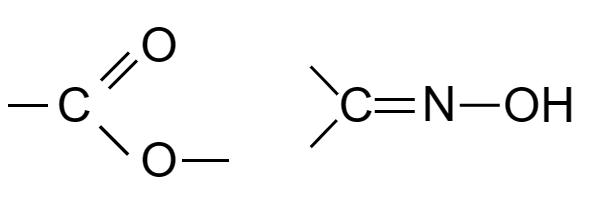}
\caption{OOD Ester (\textbf{left}) and oxime (\textbf{right}) groups present in evaluation examples. If either of these substructures is present in a molecular graph, it is removed from the training set (QM9-base) and passed to the appropriate evaluation set: QM9-OOD-Ester or QM9-OOD-Oxime. Along with oximes, all other structures with nitrogen-oxygen bond are chosen.}
\label{partition}
\end{figure}

\paragraph{Dataset creation} The experiments were conducted with QM9 dataset (\cite{wu2018moleculenet}), adequately processed for in-context learning. The goal of the experiment is to achieve improved performance on qualitatively different data by 
incorporating knowledge from several similar examples. For that, one needs to split the dataset to obtain the OOD subsets and partition each of them into contexts, in such a way that examples within each context share a 
degree of common information. We start by picking out two kinds of structures that will be later used for validation: esters and oximes (Fig. 
\ref{partition}), to obtain three partitions of data. Esters are supposed to be the less challenging dataset for generalization: each node-to-node connection is well represented in the training data, and only the emergent ester group was never seen. Oximes pose a deeper OOD 
problem: they contain a non-typical nitrogen-oxygen bond, which has never been seen in the training dataset. To further isolate such cases, we 
indiscriminately pass all other molecules with nitrogen-oxygen bonds to the oxime dataset (e.g. 
combinations of other functional groups, like hydroxylamines). QM9-Base is the part of QM9 that contains all examples except the aforementioned structures of interest. The rest are assigned to QM9-OOD-Ester and QM9-OOD-Oxime. Then, within each of the splits, we conduct a graph mining 
procedure using \textit{gspan} library (\cite{shaul2021cgspan}) and pick out frequent subgraphs. For each chosen subgraph, we randomly pick  molecular graphs that share it in order to form 10-element sequences. Details of 
the mining methodology have been outlined in the appendix.

\begin{figure}[b] 
\centering
\includegraphics[width=0.99\textwidth]{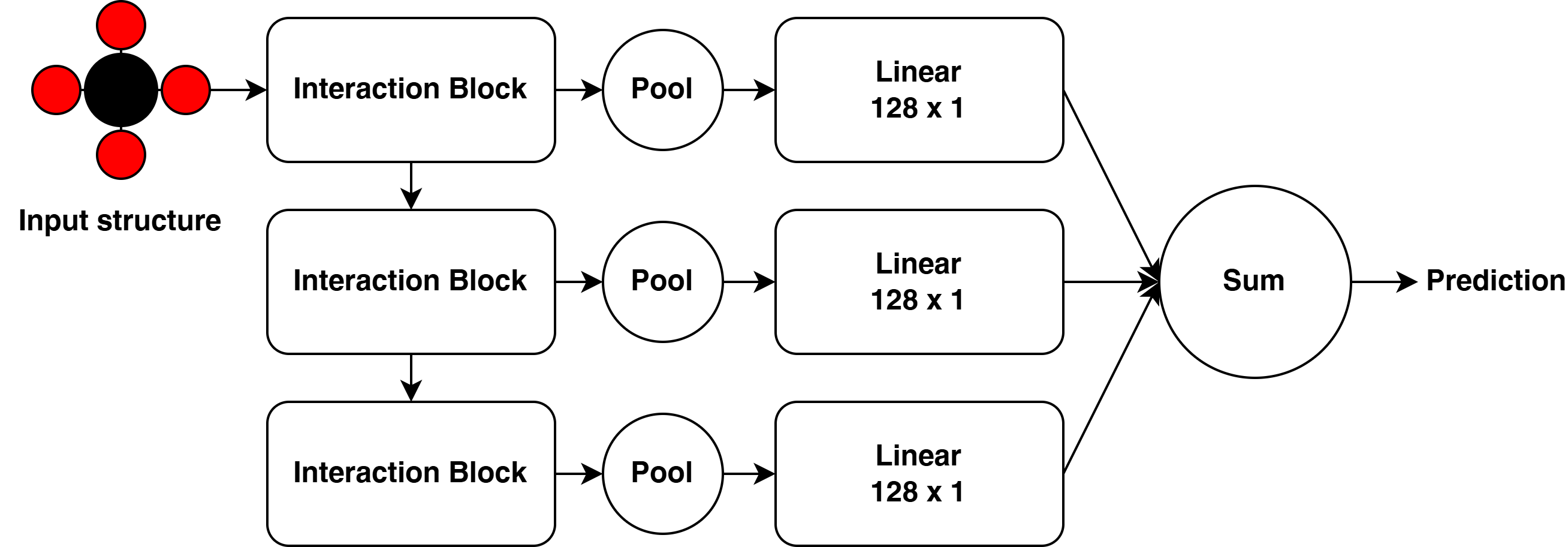}
\caption{Schematic representation of MXMNet's readout operation. We use pooled representations that describe the entire graphs, effectively replacing the linear layers shown.}
\label{mxmnet}
\end{figure}

\paragraph{Model} To process the molecular graphs, we utilize MXMNet (\cite{zhang2020molecular}) which achieves excellent performance on QM9. First, we train the model on the QM9-Base subset that does not contain examples intended as OOD. Then, the representations produced by MXMNet are used as feature vectors that describe individual structures. 
MXMNet consists of a sequence of blocks, each performing a message passing operation between adjacent edges (local message passing) as well as an additional, global message passing operation. After each block, global pooling is performed, yielding separate feature vectors, each passing a single linear layer to form partial predictions of the output label. The ultimate output is the sum of all the partial predicitons (Fig. \ref{mxmnet}). Because outputs on consecutive MXMNet blocks are used as complementary information by the model, we also concatenate them all to provide representations for GPT-2. 

\paragraph{Training task} In order to improve the estimations of the model on unknown data, we aim to utilize contextual information: 

\begin{equation}
\label{general_model}
    y = \sigma(x, C),
\end{equation}
where \( x \) is input information about the structure, \( y \) is an estimated feature, and \( C \) denotes contextual information. As the tested feature, we elect absolute-zero atomization energy \( \mathcal{U}_{0}\). We draw inspiration from \cite{garg2023transformers} as to how to redefine our regression problem as a sequence modeling task. The input passed to the model is a sequence of feature vectors that describe individual structures, followed by representations of their respective labels:

\begin{equation}
\label{simple stack}
    y_i = \sigma(x_1, y_1, x_2, y_2, (...), x_{i-1}, y_{i-1}, x_i)
\end{equation}

Sequence modeling and processing of molecular graphs are two fundamentally different tasks, with graph topology and geometry being key information, and language models acting only on a series of feature vectors. Therefore, we use GPT-2 (\(\sigma_{L}\)) on top of MXMNet (\(\sigma_{G}\)):

\begin{equation}
\label{graph stack}
    y_i = \sigma_L(\sigma_G(x_1), y_1, \sigma_G(x_2), y_2, (...), \sigma_G(x{i-1}), y_{i-1}. \sigma_G(x_i))
\end{equation}

\begin{figure}[b] 
\centering
\includegraphics[width=0.99\textwidth]{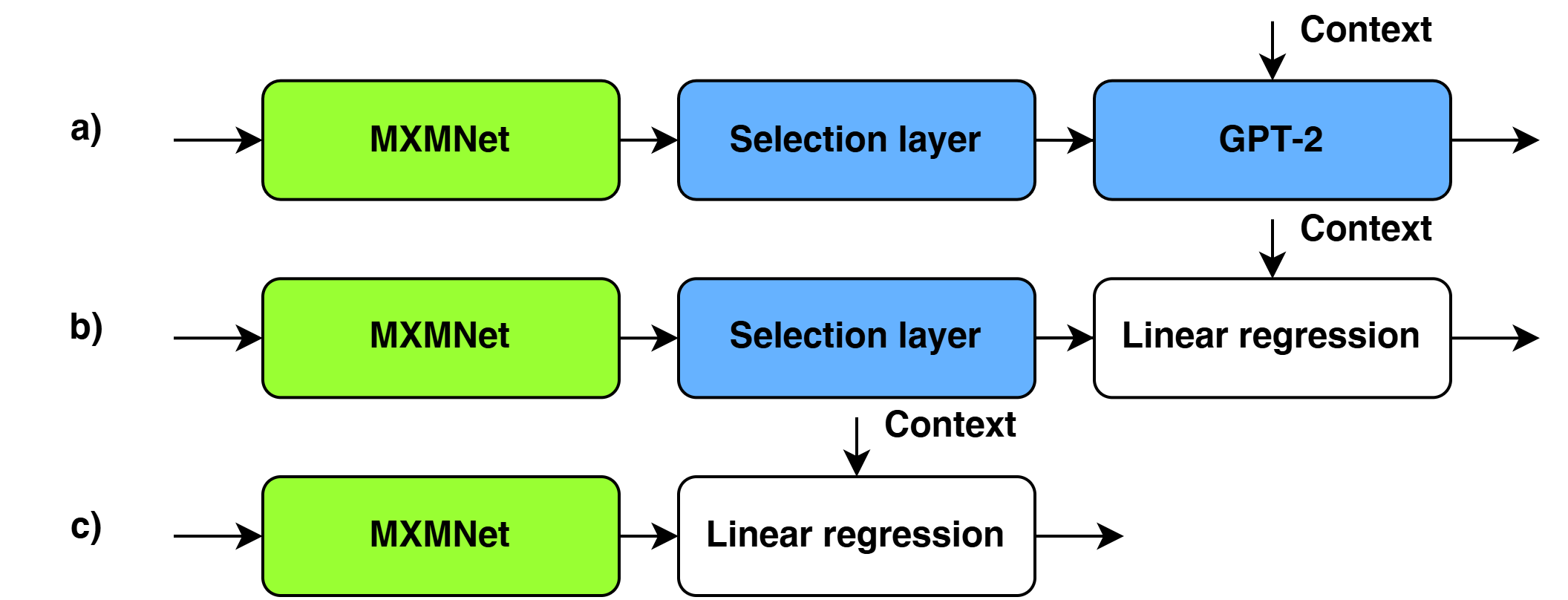}
\caption{Visualization of the in-context learning pipeline, along with two ablations considered. In green: the module trained in advance. In blue: modules fitted during in-context training. In white: a linear regression  - the last readout scheme involves no deep-learning training.}
\label{ablation_fig}
\end{figure}

The in-context pipeline is presented in Fig.(\ref{ablation_fig}). We train a single linear layer that takes the encodings produced by MXMNet and pass through a single \emph{selection} layer to match the dimensionality of GPT-2. The \emph{selection} layer is a single linear layer whose purpose is to reduce the verbose stack of MXMNet's representations to a single 128-value graph-level feature vector. Similarly, we pass the labels to a linear layer to expand them to said dimensionality. GPT-2 acts on the sequence of structure representations, each immediately followed by representations of its respective label, as in Equ. \ref{graph stack}. The model is trained to predict the labels that follow their respective structures -- the output of GPT-2 is ultimately transformed to a single value by a linear layer. When the last token is instead the representation of a label, the following prediction would be a new structure -- naturally, those outputs of GPT-2 are never used and ignored when calculating the loss. To further evaluate the final model, we performed an ablation study, explained in detail in the "Experiments" section.

\paragraph{Training curriculum}
In the sequence-modeling approach, GPT-2 can be required to accurately predict all of the labels present in a sequence. In extremal case, for the first element of the sequence, the model makes prediction solely based on the considered structure's feature vector -- with no contextual hint:
\begin{equation}
\label{general_model}
    y_1 = \sigma_L(\sigma_G(x_1)),
\end{equation}

We find that this is useful for preliminary fitting of the language model, but makes it prone to under-utilizing the contextual knowledge if that is present. We start off by considering the errors made by the model on the entire sequence. However, as the training progresses, we loosen the requirements of accurate prediction on initial, under-informed examples. We gradually decrease the importance of prediction errors on following examples in a sequence, to ultimately consider only the accuracy of the last prediction in a sequence.

\paragraph{Augmentation}
In-context examples serve as a source of information for the model, but the order of examples doesn't contain knowledge regarding the problem. To efficiently use the constructed sequences during training and augment for permutational equivariance of examples, we shuffle the order of examples within each sequence.

\section{Experiments}
\label{exp}

\paragraph{Base MXMNet.} First, we train MXMNet on QM9 dataset, without the context structure, but applying the filtering to maintain OOD datasets: the QM9-Base consists of all molecules in QM9 that do not contain ester or oxime groups, which are, in turn, assigned to QM9-Ester and QM9-Oxime, respectively. We split QM9-Base randomly into train and validation sets and observe that the accuracy on the validation part of QM9-Base is slightly better than what MXMNet authors achieved on randomly-split test set, which is a reasonable result (see Table (\ref{mxmnet_base})).


\begin{table}[h]
  \caption{MXMNet's prediction accuracy when trained on QM9-Base. Partitions obtained through random split are indicated with brackets. Asterisk (*) indicates reference results obtained by \cite{zhang2020molecular}, using the entire QM9 dataset, split randomly.}
  \label{mxmnet_base}
  \centering
  \begin{tabular}{lll}
    \toprule
    Training set & Evaluation set & Evaluation MAE [meV]           \\
    \midrule
    QM9-Base (train split) & QM9-Base (val split) & 5.68\\
     & QM9-OOD-Ester & 147.47\\
     & QM9-OOD-Oxime & 681.98 \\
    \midrule
    QM9 (train split) & QM9 (test split) & 5.90* \\
    \bottomrule
  \end{tabular}
\end{table}

\paragraph{GPT-2 on structure encodings.}
We train the in-context model based on GPT-2, along with the linear layer responsible for selecting input features from the encodings returned by MXMNet. Each example consists of 10 such encodings, each immediately followed by the representation of the respective label, for a total of 20 feature vectors in a sequence. We first train the model to predict the features of each of those structures, and gradually disregard the early examples (see Training Curriculum in Sec.(\ref{Method})). We evaluate the model on the basis of the accuracy in predicting the last label in each sequence -- the results are shown in Table (\ref{gpt2_table}). The LLM-based model performs worse on the training data. This could be attributed to a couple of reasons: the model is trained to apply a less biased approach by utilizing a much smaller portion of examples, insufficient to faithfully fit the labels. Moreover, the normalization operations performed by GPT-2 can degrade the magnitudinal information modeled in regression problems.  We observe that the model, despite being only trained on QM9-Base, is able to predict the features of remaining partitions of QM9 thanks to the in-context framework. With respect to a model that does not use in-context framework, the MAE decreased by nearly a factor of 5 on the QM9-OOD-Ester dataset, from 147.47 meV to 29.85 meV, and by a factor of 7 on the QM9-OOD-Oxime dataset, from 681.98 meV to 97.36 meV.


\begin{table}[h]
  \caption{In-context learning performance achieved through usage of pretrained MXMNet's encodings by GPT-2, on the last example in each sequence. Training sequences are pulled randomly at each epoch, while Ester and Oxime sequences are pulled once and deterministically. QM9-base was used as training set. Ester dataset was used as validation set for model selection and scheduling.}
   \label{gpt2_table}
  \centering
  \begin{tabular}{lll}
    \toprule
    Training set & Evaluation set & Last structure MAE [meV]           \\
    \midrule
    QM9-Base & QM9-Base & 21.20\\
     & QM9-OOD-Ester & \textbf{29.85}\\
     & QM9-OOD-Oxime & \textbf{97.36} \\
    \bottomrule
  \end{tabular}
\end{table}

\paragraph{Ablation study}
The usage of GPT-2 allowed for significant improvement of results on OOD data, thanks to the integration of contextual information. The question arises whether, from graph modeling perspective, the operations performed by this model are only superficial, or are able to consolidate some notrivial graph information? The language model is introduced in place of linear readout layers utilized by MXMNet, and the performance of the graph part of the pipeline does not use context to alleviate the OOD problem. We try to examine how valuable the LLM feedback could be, given this constraint.

When performing on well-known data, MXMNet provides the final output through linear combination of the embeddings used. For a sequence of OOD examples, one could use the context for linear regression, to find a new, optimal linear combination for each sequence of related examples. The in-context learning framework was constrained to utilize a limited number of examples. The initial linear layer, used for selection of features from MXMNet's verbose encodings, supposedly provides a more efficient representation. We examine whether the language model can yield better prediction than linear regression. Two alternative approached for utilizing in-context information are considered. In the first one, linear regression of features selected for inference during in-context training is used. In the other, we directly pass entire representations from MXMNet to linear regression, as shown in Fig.\ref{ablation_fig}.



Comparative results of described methods are shown in Table \ref{ablation_table}. Replacement of the linear readout in MXMNet with \emph{selection} layer followed either by LLM or linear regression provided significant improvement on QM9-Base and QM9-OOD-Ester. On the training data, the best approach was to utilize efficient representations produced by \emph{selection} layer, to then proceed with linear regression on the emergent representation. On QM9-OOD-Ester, the usage of entire in-context learning stack, including LLM, provided the best improvement. In this case, the potential of generalization through more informed inference in-context fully justified the usage of language modeling - the features of examples never seen in either part of the training were efficiently utilized in-context. Results were only slightly worse than those achieved by the language model on QM9-Base, despite the OOD setting.

\begin{table}[h]
  \caption{The performance of prediction achieved through different readout schemes for MXMNet's encodings. The results of GPT-2 are contrasted with linear regression - either after passing through the same linear layer that GPT-2 used for feature selection, or on full encodings from MXMNet.}
  \label{ablation_table}
  \centering
  \begin{tabular}{llll}
    \toprule
    Feature readout & QM9-Base & QM9-OOD-Ester & QM9-OOD-Oxime \\
    \midrule
    Selection + LLM  & 21.20 & \textbf{29.85} & 97.36 \\
     Selection + Regression & \textbf{11.45} & 38.12 & \textbf{73.04} \\
    \midrule
    Regression & 136.3 & 135.6 & 99.66 \\
    \bottomrule
  \end{tabular}
\end{table}

The difficulties posed by QM9-OOD-Oxime, namely introduction of a completely new kind of edge within the molecular graph (nitrogen-oxygen) proved to be the most disruptive to MXMNet (Table \ref{mxmnet_base}). Consequently, the representations from MXMNet did not serve as equally valuable input to the in-context learning framework. In this setting, linear regression proved the most efficient when used on the outputs of \emph{selection} layer. GPT-2 failed to provide the same quality of OOD generalization in this setting. This result suggests a boundary to in-context learning capabilities tied closely to the representations produced by a frozen graph encoder.

\section{Discussion}
\paragraph{Summary} In this work, we formulated and approached an in-context molecular property regression problem. We proposed a method that involves a joint usage of a specialized graph architecture for processing small organic molecules and a transformer to incorporate in-context information. To model the knowledge emerging from in-context examples processing, we replaced a stack of MXMNet's parallel readout layers with a context-aware language model. On the training dataset (QM9-Base), a natural upper bound of the regression accuracy is that of MXMNet, which was finely fitted to this large dataset. However, the inclusion of contextual knowledge proved valuable for predictions on structures never seen during the training of either part of the network.

The value of LLM-based in-context inference was further evaluated against an intuitive baseline of linear regression: this statistical method fits the task especially well -- the representations used had been co-adapted to learned linear combination used by MXMNet. LLM's in-context learning and linear regression are both graph-agnostic and aim to better utilize MXMNet's knowledge model in an OOD setting. GPT-2 readout achieved better results, supposedly building a more intricate hypothesis for in-context inference, backed with the knowledge about the entire training dataset.

The proposed graph mining approach is a convenient way to utilize classic benchmarks for research of in-context learning. The task of learning OOD data in-context demands both fundamental knowledge of the problem, and a collection of in-context examples. The data mining framework utilized in this paper provides flexibility in controlling which characteristics are elements of inter-contextual or intra-contextual variance of data. The structures within one context are constrained to have a common subgraph. The inclusion of any carbon-based structures on top of the subgraph allows features like size and isomery of the carbon chain, multiplicity of carbon bonds, cyclicity and aromaticity, to fall within intra-contextual variance (mining hyperparameters are listed in the appendix).

\paragraph{Limitations} The in-context framework achieves significantly worse results on the training data than MXMNet by itself. This can be attributed to the insufficient fitting of the language model readout from limited number of context examples. Another explanation is to be found in the discrepancy between the ways in which MXMNet and GPT-2 treat the magnitude of the signal in the forward pass: MXMNet, as common for physics-modeling neural networks, utilizes SiLU activation, so as not to squash the magnitude of the data; GPT-2, instead, extensively uses normalization layers to provide the basis for interaction between the elements of the sequence (i.e. to decouple the magnitude and direction of the feature vectors before considering their alignment). It can be hypothesized that this breach of physics modeling principles also impacts accuracy achievable on OOD data, and should be overcome with a dedicated architectural solution, especially for regression problems. Another limiting factor for the predictive quality of the pipeline may reside in the fact that graph-level representations of MXMNet are a simplified view of the knowledge about molecular graphs. Therefore, during in-context inference, the language model acts on a fundamentally lossy representation of the problem.


\bibliographystyle{plainnat}
\bibliography{bib.bib}

\begin{thebibliography}{45}
\providecommand{\natexlab}[1]{#1}
\providecommand{\url}[1]{\texttt{#1}}
\expandafter\ifx\csname urlstyle\endcsname\relax
  \providecommand{\doi}[1]{doi: #1}\else
  \providecommand{\doi}{doi: \begingroup \urlstyle{rm}\Url}\fi

\bibitem[Ahmad et~al.(2022)Ahmad, Simon, Chithrananda, Grand, and
  Ramsundar]{ahmad2022chemberta}
Walid Ahmad, Elana Simon, Seyone Chithrananda, Gabriel Grand, and Bharath
  Ramsundar.
\newblock Chemberta-2: Towards chemical foundation models.
\newblock \emph{arXiv preprint arXiv:2209.01712}, 2022.

\bibitem[Back et~al.(2019)Back, Yoon, Tian, Zhong, Tran, and
  Ulissi]{back2019convolutional}
Seoin Back, Junwoong Yoon, Nianhan Tian, Wen Zhong, Kevin Tran, and Zachary~W
  Ulissi.
\newblock Convolutional neural network of atomic surface structures to predict
  binding energies for high-throughput screening of catalysts.
\newblock \emph{The journal of physical chemistry letters}, 10\penalty0
  (15):\penalty0 4401--4408, 2019.

\bibitem[Batatia et~al.(2023)Batatia, Kovács, Simm, Ortner, and
  Csányi]{batatia2023mace}
Ilyes Batatia, Dávid~Péter Kovács, Gregor N.~C. Simm, Christoph Ortner, and
  Gábor Csányi.
\newblock Mace: Higher order equivariant message passing neural networks for
  fast and accurate force fields, 2023.

\bibitem[Batatia et~al.(2024)Batatia, Benner, Chiang, Elena, Kovács,
  Riebesell, Advincula, Asta, Avaylon, Baldwin, Berger, Bernstein, Bhowmik,
  Blau, Cărare, Darby, De, Pia, Deringer, Elijošius, El-Machachi, Falcioni,
  Fako, Ferrari, Genreith-Schriever, George, Goodall, Grey, Grigorev, Han,
  Handley, Heenen, Hermansson, Holm, Jaafar, Hofmann, Jakob, Jung, Kapil,
  Kaplan, Karimitari, Kermode, Kroupa, Kullgren, Kuner, Kuryla, Liepuoniute,
  Margraf, Magdău, Michaelides, Moore, Naik, Niblett, Norwood, O'Neill,
  Ortner, Persson, Reuter, Rosen, Schaaf, Schran, Shi, Sivonxay, Stenczel,
  Svahn, Sutton, Swinburne, Tilly, van~der Oord, Varga-Umbrich, Vegge,
  Vondrák, Wang, Witt, Zills, and Csányi]{batatia2024foundation}
Ilyes Batatia, Philipp Benner, Yuan Chiang, Alin~M. Elena, Dávid~P. Kovács,
  Janosh Riebesell, Xavier~R. Advincula, Mark Asta, Matthew Avaylon, William~J.
  Baldwin, Fabian Berger, Noam Bernstein, Arghya Bhowmik, Samuel~M. Blau, Vlad
  Cărare, James~P. Darby, Sandip De, Flaviano~Della Pia, Volker~L. Deringer,
  Rokas Elijošius, Zakariya El-Machachi, Fabio Falcioni, Edvin Fako, Andrea~C.
  Ferrari, Annalena Genreith-Schriever, Janine George, Rhys E.~A. Goodall,
  Clare~P. Grey, Petr Grigorev, Shuang Han, Will Handley, Hendrik~H. Heenen,
  Kersti Hermansson, Christian Holm, Jad Jaafar, Stephan Hofmann, Konstantin~S.
  Jakob, Hyunwook Jung, Venkat Kapil, Aaron~D. Kaplan, Nima Karimitari,
  James~R. Kermode, Namu Kroupa, Jolla Kullgren, Matthew~C. Kuner, Domantas
  Kuryla, Guoda Liepuoniute, Johannes~T. Margraf, Ioan-Bogdan Magdău, Angelos
  Michaelides, J.~Harry Moore, Aakash~A. Naik, Samuel~P. Niblett, Sam~Walton
  Norwood, Niamh O'Neill, Christoph Ortner, Kristin~A. Persson, Karsten Reuter,
  Andrew~S. Rosen, Lars~L. Schaaf, Christoph Schran, Benjamin~X. Shi, Eric
  Sivonxay, Tamás~K. Stenczel, Viktor Svahn, Christopher Sutton, Thomas~D.
  Swinburne, Jules Tilly, Cas van~der Oord, Eszter Varga-Umbrich, Tejs Vegge,
  Martin Vondrák, Yangshuai Wang, William~C. Witt, Fabian Zills, and Gábor
  Csányi.
\newblock A foundation model for atomistic materials chemistry, 2024.

\bibitem[Behler(2011)]{behler2011atom}
J{\"o}rg Behler.
\newblock Atom-centered symmetry functions for constructing high-dimensional
  neural network potentials.
\newblock \emph{The Journal of chemical physics}, 134\penalty0 (7), 2011.

\bibitem[Brown et~al.(2020)Brown, Mann, Ryder, Subbiah, Kaplan, Dhariwal,
  Neelakantan, Shyam, Sastry, Askell, Agarwal, Herbert-Voss, Krueger, Henighan,
  Child, Ramesh, Ziegler, Wu, Winter, Hesse, Chen, Sigler, Litwin, Gray, Chess,
  Clark, Berner, McCandlish, Radford, Sutskever, and Amodei]{brown2020language}
Tom~B. Brown, Benjamin Mann, Nick Ryder, Melanie Subbiah, Jared Kaplan,
  Prafulla Dhariwal, Arvind Neelakantan, Pranav Shyam, Girish Sastry, Amanda
  Askell, Sandhini Agarwal, Ariel Herbert-Voss, Gretchen Krueger, Tom Henighan,
  Rewon Child, Aditya Ramesh, Daniel~M. Ziegler, Jeffrey Wu, Clemens Winter,
  Christopher Hesse, Mark Chen, Eric Sigler, Mateusz Litwin, Scott Gray,
  Benjamin Chess, Jack Clark, Christopher Berner, Sam McCandlish, Alec Radford,
  Ilya Sutskever, and Dario Amodei.
\newblock Language models are few-shot learners, 2020.

\bibitem[Chan et~al.(2022)Chan, Santoro, Lampinen, Wang, Singh, Richemond,
  McClelland, and Hill]{chan2022data}
Stephanie Chan, Adam Santoro, Andrew Lampinen, Jane Wang, Aaditya Singh, Pierre
  Richemond, James McClelland, and Felix Hill.
\newblock Data distributional properties drive emergent in-context learning in
  transformers.
\newblock \emph{Advances in Neural Information Processing Systems},
  35:\penalty0 18878--18891, 2022.

\bibitem[Chen et~al.(2019)Chen, Ye, Zuo, Zheng, and Ong]{chen2019graph}
Chi Chen, Weike Ye, Yunxing Zuo, Chen Zheng, and Shyue~Ping Ong.
\newblock Graph networks as a universal machine learning framework for
  molecules and crystals.
\newblock \emph{Chemistry of Materials}, 31\penalty0 (9):\penalty0 3564--3572,
  2019.

\bibitem[Chithrananda et~al.(2020)Chithrananda, Grand, and
  Ramsundar]{chithrananda2020chemberta}
Seyone Chithrananda, Gabriel Grand, and Bharath Ramsundar.
\newblock Chemberta: large-scale self-supervised pretraining for molecular
  property prediction.
\newblock \emph{arXiv preprint arXiv:2010.09885}, 2020.

\bibitem[Choudhary and DeCost(2021)]{choudhary2021atomistic}
Kamal Choudhary and Brian DeCost.
\newblock Atomistic line graph neural network for improved materials property
  predictions.
\newblock \emph{npj Computational Materials}, 7\penalty0 (1):\penalty0 185,
  2021.

\bibitem[De et~al.(2016)De, Bart{\'o}k, Cs{\'a}nyi, and
  Ceriotti]{de2016comparing}
Sandip De, Albert~P Bart{\'o}k, G{\'a}bor Cs{\'a}nyi, and Michele Ceriotti.
\newblock Comparing molecules and solids across structural and alchemical
  space.
\newblock \emph{Physical Chemistry Chemical Physics}, 18\penalty0
  (20):\penalty0 13754--13769, 2016.

\bibitem[Duvenaud et~al.(2015)Duvenaud, Maclaurin, Iparraguirre, Bombarell,
  Hirzel, Aspuru-Guzik, and Adams]{duvenaud2015convolutional}
David~K Duvenaud, Dougal Maclaurin, Jorge Iparraguirre, Rafael Bombarell,
  Timothy Hirzel, Al{\'a}n Aspuru-Guzik, and Ryan~P Adams.
\newblock Convolutional networks on graphs for learning molecular fingerprints.
\newblock \emph{Advances in neural information processing systems}, 28, 2015.

\bibitem[Edwards et~al.(2023)Edwards, Naik, Khot, Burke, Ji, and
  Hope]{edwards2023synergpt}
Carl~N Edwards, Aakanksha Naik, Tushar Khot, Martin~D Burke, Heng Ji, and Tom
  Hope.
\newblock Synergpt: In-context learning for personalized drug synergy
  prediction and drug design.
\newblock \emph{bioRxiv}, pages 2023--07, 2023.

\bibitem[Garg et~al.(2023)Garg, Tsipras, Liang, and
  Valiant]{garg2023transformers}
Shivam Garg, Dimitris Tsipras, Percy Liang, and Gregory Valiant.
\newblock What can transformers learn in-context? a case study of simple
  function classes.
\newblock 2023.

\bibitem[Gasteiger et~al.(2020)Gasteiger, Gro{\ss}, and
  G{\"u}nnemann]{gasteiger2020directional}
Johannes Gasteiger, Janek Gro{\ss}, and Stephan G{\"u}nnemann.
\newblock Directional message passing for molecular graphs.
\newblock \emph{arXiv preprint arXiv:2003.03123}, 2020.

\bibitem[Gasteiger et~al.(2021)Gasteiger, Becker, and
  G{\"u}nnemann]{gasteiger2021gemnet}
Johannes Gasteiger, Florian Becker, and Stephan G{\"u}nnemann.
\newblock Gemnet: Universal directional graph neural networks for molecules.
\newblock \emph{Advances in Neural Information Processing Systems},
  34:\penalty0 6790--6802, 2021.

\bibitem[Guo et~al.(2023{\natexlab{a}})Guo, Guo, Nan, Liang, Guo, Chawla,
  Wiest, and Zhang]{guo2023large}
Taicheng Guo, Kehan Guo, Bozhao Nan, Zhenwen Liang, Zhichun Guo, Nitesh~V.
  Chawla, Olaf Wiest, and Xiangliang Zhang.
\newblock What can large language models do in chemistry? a comprehensive
  benchmark on eight tasks, 2023{\natexlab{a}}.

\bibitem[Guo et~al.(2023{\natexlab{b}})Guo, Nan, Liang, Guo, Chawla, Wiest,
  Zhang, et~al.]{guo2023can}
Taicheng Guo, Bozhao Nan, Zhenwen Liang, Zhichun Guo, Nitesh Chawla, Olaf
  Wiest, Xiangliang Zhang, et~al.
\newblock What can large language models do in chemistry? a comprehensive
  benchmark on eight tasks.
\newblock \emph{Advances in Neural Information Processing Systems},
  36:\penalty0 59662--59688, 2023{\natexlab{b}}.

\bibitem[Huang et~al.(2024)Huang, Ren, Chen, Kr{\v{z}}manc, Zeng, Liang, and
  Leskovec]{huang2024prodigy}
Qian Huang, Hongyu Ren, Peng Chen, Gregor Kr{\v{z}}manc, Daniel Zeng, Percy~S
  Liang, and Jure Leskovec.
\newblock Prodigy: Enabling in-context learning over graphs.
\newblock \emph{Advances in Neural Information Processing Systems}, 36, 2024.

\bibitem[Isayev et~al.(2017)Isayev, Oses, Toher, Gossett, Curtarolo, and
  Tropsha]{isayev2017universal}
Olexandr Isayev, Corey Oses, Cormac Toher, Eric Gossett, Stefano Curtarolo, and
  Alexander Tropsha.
\newblock Universal fragment descriptors for predicting properties of inorganic
  crystals.
\newblock \emph{Nature communications}, 8\penalty0 (1):\penalty0 15679, 2017.

\bibitem[Kipf and Welling(2016)]{kipf2016semi}
Thomas~N Kipf and Max Welling.
\newblock Semi-supervised classification with graph convolutional networks.
\newblock \emph{arXiv preprint arXiv:1609.02907}, 2016.

\bibitem[Lampinen et~al.(2022)Lampinen, Dasgupta, Chan, Matthewson, Tessler,
  Creswell, McClelland, Wang, and Hill]{lampinen2022can}
Andrew~K Lampinen, Ishita Dasgupta, Stephanie~CY Chan, Kory Matthewson,
  Michael~Henry Tessler, Antonia Creswell, James~L McClelland, Jane~X Wang, and
  Felix Hill.
\newblock Can language models learn from explanations in context?
\newblock \emph{arXiv preprint arXiv:2204.02329}, 2022.

\bibitem[Li and Jiang(2021)]{li2021mol}
Juncai Li and Xiaofei Jiang.
\newblock Mol-bert: an effective molecular representation with bert for
  molecular property prediction.
\newblock \emph{Wireless Communications and Mobile Computing}, 2021:\penalty0
  1--7, 2021.

\bibitem[Liu et~al.(2021)Liu, Shen, Zhang, Dolan, Carin, and
  Chen]{liu2021makes}
Jiachang Liu, Dinghan Shen, Yizhe Zhang, Bill Dolan, Lawrence Carin, and Weizhu
  Chen.
\newblock What makes good in-context examples for gpt-$3 $?
\newblock \emph{arXiv preprint arXiv:2101.06804}, 2021.

\bibitem[Merchant et~al.(2023)Merchant, Batzner, Schoenholz, Aykol, Cheon, and
  Cubuk]{Merchant2023}
Amil Merchant, Simon Batzner, Samuel~S. Schoenholz, Muratahan Aykol, Gowoon
  Cheon, and Ekin~Dogus Cubuk.
\newblock Scaling deep learning for materials discovery.
\newblock \emph{Nature}, 624\penalty0 (7990):\penalty0 80--85, Dec 2023.
\newblock ISSN 1476-4687.
\newblock \doi{10.1038/s41586-023-06735-9}.
\newblock URL \url{https://doi.org/10.1038/s41586-023-06735-9}.

\bibitem[Min et~al.(2021{\natexlab{a}})Min, Lewis, Hajishirzi, and
  Zettlemoyer]{min2021noisy}
Sewon Min, Mike Lewis, Hannaneh Hajishirzi, and Luke Zettlemoyer.
\newblock Noisy channel language model prompting for few-shot text
  classification.
\newblock \emph{arXiv preprint arXiv:2108.04106}, 2021{\natexlab{a}}.

\bibitem[Min et~al.(2021{\natexlab{b}})Min, Lewis, Zettlemoyer, and
  Hajishirzi]{min2021metaicl}
Sewon Min, Mike Lewis, Luke Zettlemoyer, and Hannaneh Hajishirzi.
\newblock Metaicl: Learning to learn in context.
\newblock \emph{arXiv preprint arXiv:2110.15943}, 2021{\natexlab{b}}.

\bibitem[Olsson et~al.(2022)Olsson, Elhage, Nanda, Joseph, DasSarma, Henighan,
  Mann, Askell, Bai, Chen, et~al.]{olsson2022context}
Catherine Olsson, Nelson Elhage, Neel Nanda, Nicholas Joseph, Nova DasSarma,
  Tom Henighan, Ben Mann, Amanda Askell, Yuntao Bai, Anna Chen, et~al.
\newblock In-context learning and induction heads.
\newblock \emph{arXiv preprint arXiv:2209.11895}, 2022.

\bibitem[Ramakrishna et~al.(2019)Ramakrishna, Zhang, Lu, Qian, Low, Yune, Tan,
  Bressan, Sanvito, and Kalidindi]{ramakrishna2019materials}
Seeram Ramakrishna, Tong-Yi Zhang, Wen-Cong Lu, Quan Qian, Jonathan Sze~Choong
  Low, Jeremy Heiarii~Ronald Yune, Daren Zong~Loong Tan, St{\'e}phane Bressan,
  Stefano Sanvito, and Surya~R Kalidindi.
\newblock Materials informatics.
\newblock \emph{Journal of Intelligent Manufacturing}, 30:\penalty0 2307--2326,
  2019.

\bibitem[Ramos et~al.(2023)Ramos, Michtavy, Porosoff, and
  White]{ramos2023bayesian}
Mayk~Caldas Ramos, Shane~S. Michtavy, Marc~D. Porosoff, and Andrew~D. White.
\newblock Bayesian optimization of catalysts with in-context learning, 2023.

\bibitem[Rong(2021)]{rong2021extrapolating}
Frieda Rong.
\newblock Extrapolating to unnatural language processing with gpt-3’s
  in-context learning: The good, the bad, and the mysterious, 2021.

\bibitem[Ryczko et~al.(2018)Ryczko, Mills, Luchak, Homenick, and
  Tamblyn]{ryczko2018convolutional}
Kevin Ryczko, Kyle Mills, Iryna Luchak, Christa Homenick, and Isaac Tamblyn.
\newblock Convolutional neural networks for atomistic systems.
\newblock \emph{Computational Materials Science}, 149:\penalty0 134--142, 2018.

\bibitem[Scarselli et~al.(2008)Scarselli, Gori, Tsoi, Hagenbuchner, and
  Monfardini]{scarselli2008graph}
Franco Scarselli, Marco Gori, Ah~Chung Tsoi, Markus Hagenbuchner, and Gabriele
  Monfardini.
\newblock The graph neural network model.
\newblock \emph{IEEE transactions on neural networks}, 20\penalty0
  (1):\penalty0 61--80, 2008.

\bibitem[Sch{\"u}tt et~al.(2017)Sch{\"u}tt, Kindermans, Sauceda~Felix, Chmiela,
  Tkatchenko, and M{\"u}ller]{schutt2017schnet}
Kristof Sch{\"u}tt, Pieter-Jan Kindermans, Huziel~Enoc Sauceda~Felix, Stefan
  Chmiela, Alexandre Tkatchenko, and Klaus-Robert M{\"u}ller.
\newblock Schnet: A continuous-filter convolutional neural network for modeling
  quantum interactions.
\newblock \emph{Advances in neural information processing systems}, 30, 2017.

\bibitem[Shaul and Naaz(2021)]{shaul2021cgspan}
Zevin Shaul and Sheikh Naaz.
\newblock cgspan: Closed graph-based substructure pattern mining, 2021.

\bibitem[Vaswani et~al.(2017)Vaswani, Shazeer, Parmar, Uszkoreit, Jones, Gomez,
  Kaiser, and Polosukhin]{vaswani2017attention}
Ashish Vaswani, Noam Shazeer, Niki Parmar, Jakob Uszkoreit, Llion Jones,
  Aidan~N Gomez, {\L}ukasz Kaiser, and Illia Polosukhin.
\newblock Attention is all you need.
\newblock \emph{Advances in neural information processing systems}, 30, 2017.

\bibitem[Wang et~al.(2019)Wang, Guo, Wang, Sun, and Huang]{wang2019smiles}
Sheng Wang, Yuzhi Guo, Yuhong Wang, Hongmao Sun, and Junzhou Huang.
\newblock Smiles-bert: large scale unsupervised pre-training for molecular
  property prediction.
\newblock In \emph{Proceedings of the 10th ACM international conference on
  bioinformatics, computational biology and health informatics}, pages
  429--436, 2019.

\bibitem[Ward and Wolverton(2017)]{ward2017atomistic}
Logan Ward and Chris Wolverton.
\newblock Atomistic calculations and materials informatics: A review.
\newblock \emph{Current Opinion in Solid State and Materials Science},
  21\penalty0 (3):\penalty0 167--176, 2017.

\bibitem[Wu et~al.(2018)Wu, Ramsundar, Feinberg, Gomes, Geniesse, Pappu,
  Leswing, and Pande]{wu2018moleculenet}
Zhenqin Wu, Bharath Ramsundar, Evan~N. Feinberg, Joseph Gomes, Caleb Geniesse,
  Aneesh~S. Pappu, Karl Leswing, and Vijay Pande.
\newblock Moleculenet: A benchmark for molecular machine learning, 2018.

\bibitem[Xie et~al.(2021)Xie, Raghunathan, Liang, and Ma]{xie2021explanation}
Sang~Michael Xie, Aditi Raghunathan, Percy Liang, and Tengyu Ma.
\newblock An explanation of in-context learning as implicit bayesian inference.
\newblock \emph{arXiv preprint arXiv:2111.02080}, 2021.

\bibitem[Xie and Grossman(2018)]{xie2018crystal}
Tian Xie and Jeffrey~C Grossman.
\newblock Crystal graph convolutional neural networks for an accurate and
  interpretable prediction of material properties.
\newblock \emph{Physical review letters}, 120\penalty0 (14):\penalty0 145301,
  2018.

\bibitem[Xue et~al.(2016)Xue, Balachandran, Hogden, Theiler, Xue, and
  Lookman]{xue2016accelerated}
Dezhen Xue, Prasanna~V Balachandran, John Hogden, James Theiler, Deqing Xue,
  and Turab Lookman.
\newblock Accelerated search for materials with targeted properties by adaptive
  design.
\newblock \emph{Nature communications}, 7\penalty0 (1):\penalty0 1--9, 2016.

\bibitem[Yang et~al.(2024)Yang, Hu, Zhou, Liu, Shi, Li, Li, Chen, Chen, Zeni,
  Horton, Pinsler, Fowler, Zügner, Xie, Smith, Sun, Wang, Kong, Liu, Hao, and
  Lu]{yang2024mattersim}
Han Yang, Chenxi Hu, Yichi Zhou, Xixian Liu, Yu~Shi, Jielan Li, Guanzhi Li,
  Zekun Chen, Shuizhou Chen, Claudio Zeni, Matthew Horton, Robert Pinsler,
  Andrew Fowler, Daniel Zügner, Tian Xie, Jake Smith, Lixin Sun, Qian Wang,
  Lingyu Kong, Chang Liu, Hongxia Hao, and Ziheng Lu.
\newblock Mattersim: A deep learning atomistic model across elements,
  temperatures and pressures, 2024.

\bibitem[Zhang et~al.(2020)Zhang, Liu, and Xie]{zhang2020molecular}
Shuo Zhang, Yang Liu, and Lei Xie.
\newblock Molecular mechanics-driven graph neural network with multiplex graph
  for molecular structures.
\newblock \emph{arXiv preprint arXiv:2011.07457}, 2020.

\bibitem[Zheng et~al.(2018)Zheng, Zheng, and Zhang]{zheng2018machine}
Xiaolong Zheng, Peng Zheng, and Rui-Zhi Zhang.
\newblock Machine learning material properties from the periodic table using
  convolutional neural networks.
\newblock \emph{Chemical science}, 9\penalty0 (44):\penalty0 8426--8432, 2018.

\end{thebibliography}


\newpage

\appendix

\section{Hyperparameters}

\paragraph{Model architecture}
We use MXMNet with 6 interaction blocks and a latent dimension of 128 channels. GPT-2 model has 12 blocks and a latent dimension of 128 channels.
Before passing the structure and label representations as a sequence, we transform them to have the same dimensionality. This structure encoding layer has 748 input channels (takes in 6 vectors, 128 scalars each) and 128 output channels. Label encoding layer has 1 input channel and 128 output channels. The output of GPT-2 is transformer by a layer with 128 input channels and 1 output channel.

\paragraph{Training}
MXMNet was trained for 900 epochs with learning rate of $10^{-4}$, and a warmup schedule. The loss criterion was mean absolute error. No scaling was applied. In evaluation, Exponential Moving Average (EMA) of model's parameters was used (cite).Batch size was 64. The training took 2 days on NVIDIA RTX 4070 GPU.

GPT-2 was trained for 2500 epochs. Learning rate was 0.001, reduced by 10 times whenever the training plateaued for 100 epochs, down to /( 10\^-5/). Batch size was 16, meaning 16 sequences, 10 examples each, per step. The loss criterion was mean squared error and both input variables (encodings by MXMNet) and labels were standardized. Loss function curriculum was defined by 2 parameters: the last example in a sequence which is ignored by the loss function and the first one fully considered, with importance of in-between examples being gradually increased. Both incremented every 600 steps (roughly 20 epochs).The index of first fully considered example started at 0 - that means, after first 20 epochs, the importance of the first example in a sequence started decreasing. The index of the last ignored example started at -5, meaning the schedule for fully ignoring loss on initial elements started after 5 20-epoch intervals. Ultimately, the accuracy of last, fully informed prediction made entirety of the loss. The training took 8 hours on NVIDIA RTX 4070 GPU.

\paragraph{Graph mining}
In the graph mining, a subgraph is defined by the types of atoms and bonds between them. We search for subgraphs of at least 2 heavy (non-hydrogen) atoms, with the additional requirement that at least one of them is non-carbon, and no more than 2 are carbon atoms. As a result, some patterns consist of just a simple functional groups, while others contain a more complex pattern, with fewer supporting examples. Individual examples that share this subgraph can have up to 6 extra carbon atoms, while other heavy atoms are not allowed. We randomly draw structures that share a common pattern to form 10-example contexts, without return. To not overuse the numerous structures from simple patterns, for each pattern we sample up to 15 contexts.

\end{document}